\documentclass{article}
\usepackage[accepted]{icml2012}
\usepackage{natbib}
\usepackage[scriptsize]{subfigure}

\usepackage[tight]{bibhacks}
\usepackage{graphicx}
\usepackage{color}
\usepackage{url}
\usepackage{amsmath}
\usepackage{amssymb}
\usepackage[mathscr]{eucal}
\usepackage{multirow}

\def\sigm{{\rm sigm}}

\def\reals{{\mathbb R}}

\def\D{{\mathbf D}}
\def\U{{\mathbf U}}

\def\v{{\mathbf v}}
\def\h{{\mathbf h}}
\def\W{{\mathbf W}}

\def\b{{\mathbf b}}
\def\c{{\mathbf c}}

\def\e{{\mathbf e}}

\newcommand{\E}{\mathbb{E}} 

\icmltitlerunning{Training Restricted Boltzmann Machines on Word Observations}

\begin{document}

\twocolumn[

\icmltitle{
Training Restricted Boltzmann Machines on Word Observations
}

\icmlauthor{George E.\ Dahl}{gdahl@cs.toronto.edu}
\icmladdress{Department of Computer Science, University of Toronto, Toronto, ON, Canada} 
\icmlauthor{Ryan P.\ Adams}{rpa@seas.harvard.edu}
\icmladdress{School of Engineering and Applied Sciences, Harvard University, Cambridge, MA, USA} 
\icmlauthor{Hugo Larochelle}{hugo.larochelle@usherbrooke.ca}
\icmladdress{D\'epartement d'informatique, Universit\'e de Sherbrooke, Sherbrooke, QC, Canada} 

\icmlkeywords{restricted Boltzmann machines, deep learning, neural networks, semi-supervised learning}
\vskip 0.3in
]

\begin{abstract}
The restricted Boltzmann machine (RBM) is a flexible model for
complex data. However, using RBMs for high-dimensional multinomial
observations poses significant computational
difficulties. In natural language processing applications, words are
naturally modeled by $K$-ary discrete distributions, where $K$ is
determined by the vocabulary size and can easily be in the hundred 
thousands. The conventional approach to training RBMs on word
observations is limited because it requires sampling the states of
$K$-way softmax visible units during block Gibbs updates, an operation
that takes time linear in $K$. In this work, we address this issue with
a more general class of Markov chain Monte Carlo operators
on the visible units, yielding updates with computational complexity
independent of $K$. We demonstrate the success of our approach by
training RBMs on hundreds of millions of word $n$-grams using larger
vocabularies than previously feasible with RBMs and by using the learned features
to improve performance on chunking and sentiment classification tasks,
achieving state-of-the-art results on the latter.

\end{abstract}

\section{Introduction}

The breadth of applications for the restricted Boltzmann machine
(RBM)~\citep{Smolensky86,FreundY1991} has expanded rapidly in recent
years.  For example, RBMs have been used to model image
patches~\citep{ranzato2010factored}, text documents as
bags of words~\citep{Salakhutdinov-NIPS2010-softmax}, and movie
ratings~\citep{SalakhutdinovR2007b-small}, among other data.  Although
RBMs were originally developed for binary observations, they have been
generalized to several other types of data, including integer- and
real-valued observations~\citep{WellingNIPS17-small}.

However, one type of data that is not well supported by the RBM is
word observations from a large vocabulary (e.g., 100,000 words). The
issue is not one of representing such observations in the RBM
framework: so-called \emph{softmax}
units~\citep{Salakhutdinov-NIPS2010-softmax} are the natural choice
for modeling words. The issue is that manipulating distributions over the
states of such units is expensive even for intermediate vocabulary
sizes and becomes impractical for vocabulary sizes in the
hundred thousands --- a typical situation for NLP problems.  For
example, with a vocabulary of 100,000 words, modeling $n$-gram windows
of size ${n=5}$ is comparable in scale to training an RBM on binary
vector observations of dimension 500,000 (i.e., more dimensions than a
${700 \times  700}$ pixel image).  This
scalability issue has been a primary obstacle to using the RBM
for natural language processing.


In this work, we directly address the scalability issues associated
with large softmax visible units in RBMs. We describe a learning rule
with a computational complexity independent of the number of
visible units. We obtain this rule by replacing the Gibbs sampling
transition kernel over the visible units with carefully implemented
Metropolis--Hastings transitions.  By training RBMs in this way on hundreds of millions
of word windows, they learn representations capturing
meaningful syntactic and semantic properties of words. Our learned
word representations provide benefits on a chunking task competitive
with other methods of inducing word representations and our learned
$n$-gram features yield even larger performance gains.  Finally, we
also show how similarly extracted $n$-gram representations can be used
to obtain state-of-the-art performance on a sentiment classification
benchmark.



\section{Restricted Boltzmann Machines}
\label{sec:rbm}

We first describe the restricted Boltzmann machine for binary
observations, which provides the basis for other data types.  An RBM
defines a distribution over a binary visible vector $\v$ of
dimensionality $V$ and a layer $\h$ of $H$ binary hidden units
through an energy
\begin{align}
  E(\v,\h) = -\b^\top \v - \c^\top \h - \h^\top \W \v. \label{eqn:energy} 
\end{align}
This energy is parameterized by bias vectors ${\b\in\reals^V}$ and ${\c\in\reals^H}$ and weight matrix~${\W\in\reals^{H\times V}}$, and 
is converted into a probability distribution via
\begin{align}
  p(\v,\h) & = \exp \left( -E(\v,\h) \right) \, / \,Z \label{eqn:boltz1} \\
  Z & = \sum_{\v',\h'}\exp(-E(\v',\h'))~.\label{eqn:boltz2}
\end{align}
This yields simple conditional distributions:
\begin{flalign}
  p(\h|\v) &= \prod_j p(h_j| \v) &
  p(\v|\h) &= \prod_i p(v_i| \h)\label{eqn:cond1}
\end{flalign}
\vspace{-0.7cm}%
\begin{align}
  p(h_j=1|\v) &= \sigm(c_j + \sum_i W_{ji} v_i)\label{eqn:hidcond2} \\
  p(v_i=1|\h) &= \sigm(b_i + \sum_j W_{ji} h_j),\label{eqn:viscond2}
\end{align}
where $\sigm(z) = 1 / (1 + e^{-z})$, which allow for efficient Gibbs sampling of each layer
given the other layer.

We train an RBM from~$T$ visible data vectors~$\{\v_t\}^T_{t=1}$ by
minimizing the scaled negative (in practice, penalized) log likelihood
of the parameters ${\theta=(\b,\c,\W)}$:
\begin{flalign}
  \theta_{\sf{MLE}} &= \underset{\theta}{\operatorname{argmin}} \;{\cal L(\theta)}  &
  {\cal L(\theta)} & = \frac{1}{T} \sum_{t} \ell(\v_t;\theta)
\end{flalign}
\vspace{-0.6cm}%
\begin{align}
  \ell(\v_t;\theta) & = -\log p(\v_t) = -\log \sum_{\h} p(\v_t,\h).
\end{align}
The gradient of the objective with respect to $\theta$
\begin{align*}
  \frac{\partial {\cal L}(\theta) }{\partial \theta} &=
  \frac{1}{T}\sum_t \E_{\h | \v_t}\left[\frac{\partial E(\v_t,\h)}{\partial \theta}\right] - \E_{\v,\h}\left[\frac{ \partial E(\v,\h)}{\partial \theta}\right]
\end{align*}
is intractable to compute because of the exponentially many terms in
the sum over joint configurations of the visible and hidden units in
the second expectation.

Fortunately, for a given~$\theta$, we can approximate this gradient by
replacing the second expectation with a Monte Carlo estimate based
on a set of~$M$ samples ${{\cal N} = \{\tilde{\v}_m\}}$ from the RBM's
distribution:
\begin{align}
  \E_{\v,\h}\left[\frac{ \partial E(\v,\h)}{\partial \theta}\right]
  &\approx
  \frac{1}{M}\!\!\sum_{\tilde{\v}_m\in{\cal N}}\!\!
  \E_{\h|\tilde{\v}_m}\!\!\left[\frac{ \partial E(\tilde{\v}_m,\h)}{\partial \theta}\right]~.\label{eqn:approx_gradient}
\end{align}
The samples $\{\tilde{\v}_m\}$ are often referred to as ``negative
samples'' as they counterbalance the gradient due to the observed, or
``positive'' data. To obtain these samples, we maintain $M$
parallel Markov chains throughout learning and update them using Gibbs sampling between parameter updates.

\begin{figure}
\begin{center}
\includegraphics[width=0.44\textwidth]{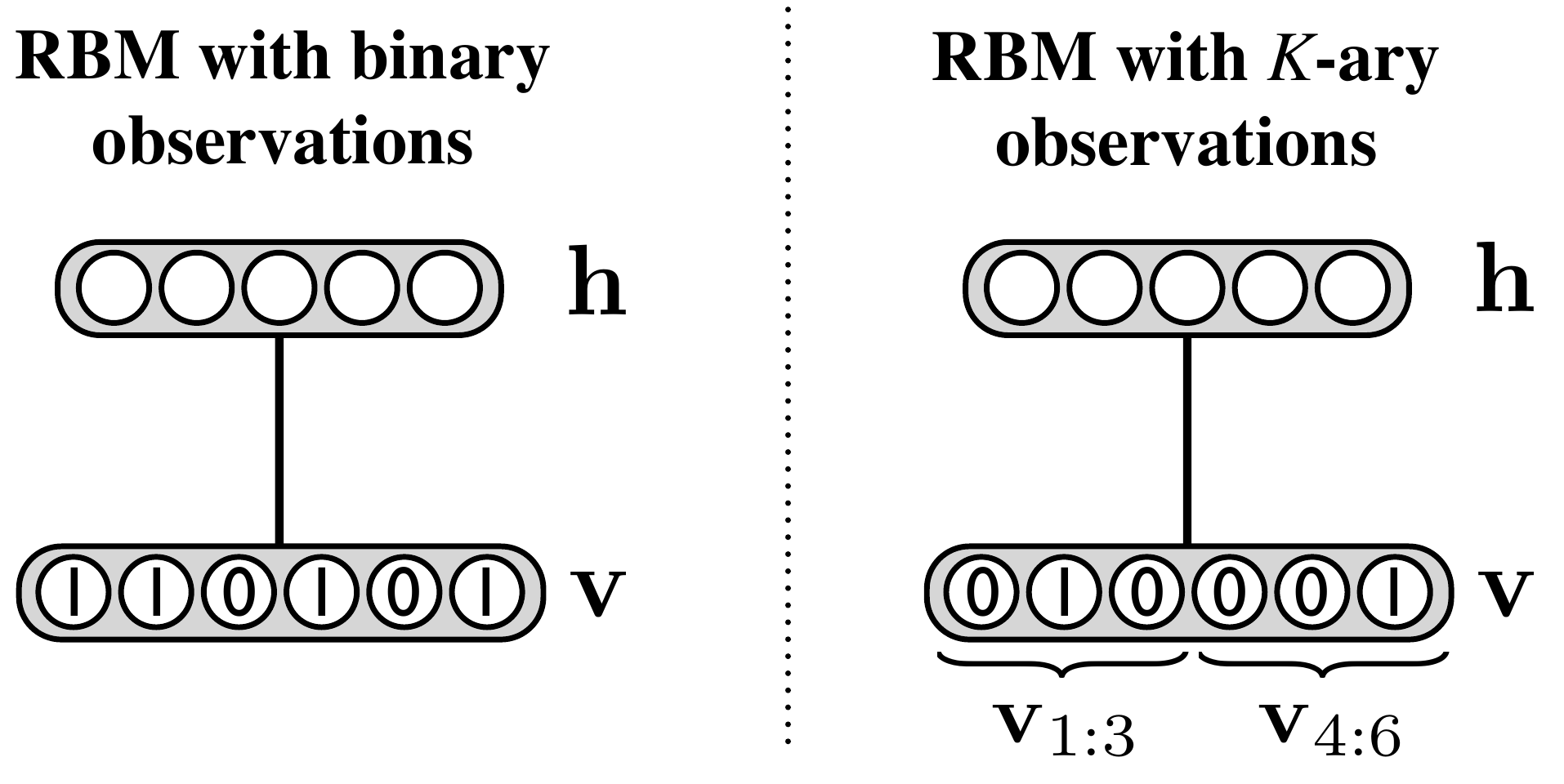}
\end{center}
\vspace*{-0.55cm}
\caption{Illustration of an RBM with binary observations (left) and
  $K$-ary observations, for $n=2$ and $K=3$, i.e.\ a pair of $3$-ary
  observations (right).}
\label{fig:k_ary_rbm}
\vspace*{-0.52cm}
\end{figure}

Learning with the Monte Carlo estimator alternates
between two steps: 1)~Using the current parameters~$\theta$, simulate
a fews steps of the Markov chain on the~$M$ negative samples using
Eqs.~(\ref{eqn:cond1})-(\ref{eqn:viscond2}); and 2)~Using the negative
samples, along with a mini-batch (subset) of the positive data,
compute the gradient in Eq.~(\ref{eqn:approx_gradient}) and update
the parameters.  This procedure belongs to the general class of
Robbins-Monro stochastic approximation algorithms~\citep{YounesL1989}.
Under mild conditions, which include the requirement that the
Markov chain operators leave~$p(\v,\h\,|\,\theta)$ invariant, this
procedure will converge to a stable point of ${\cal L}(\theta)$.


For $K$-ary observations --- observations belonging to a finite set
of~$K$ discrete outcomes --- we can use the same energy function as in
Eq.~(\ref{eqn:energy}) for the binary RBM by encoding each observation
in a ``one-hot'' representation and concatenating the representations
of all observations to construct $\v$.  In other words, for $n$
separate $K$-ary observations, the visible units $\v$ will be
partitioned into $n$ groups of $K$ binary units, with the constraint
that each partition can only contain a single non-zero entry. Using
the notation $\v_{a:b}$ to refer to the subvector of elements
from index $a$ to index $b$, the $i^{\rm th}$ observation will then be
encoded by the group of visible units $\v_{(i-1)K+1 : iK}$.  The
one-hot encoding is enforced by constraining each group of units
to contain only a single 1-valued unit, the
others being set to 0.  The difference between RBMs with binary and
$K$-ary observations is illustrated in Figure~\ref{fig:k_ary_rbm}.

To simplify notation, we refer to the $i^{\rm th}$ group of visible units as
${\v^{(i)} \!=\! \v_{(i-1)K+1 : iK}}$.  Similarly, we will refer to the biases
and weights associated with those units as ${\b^{(i)} = \b_{(i-1)K+1: iK}}$
and ${\W^{(i)} = \W_{\cdot, (i-1)K+1: iK}}$. We will also denote with $\e_k$
the one-hot vector with its $k^{\rm th}$ component set to 1.

The conditional distribution over the visible layer is
\begin{align}
  p(\v|\h) &= \prod_{i=1}^n p(\v^{(i)} | \h) \label{eqn:karyviscond}\\
  p(\v^{(i)} \!=\! \e_k |\h) &= \frac{\exp(\b^{(i)^\top} \e_k + \h^\top \W^{(i)} \e_k)}{\sum_{k'} \exp(\b^{(i)i^\top} \e_{k'} \!+\! \h^\top \W^{(i)} \e_{k'})}.\notag
\end{align}
Each group $\v^{(i)}$ has a multinomial distribution given the
hidden layer. Because the multinomial probabilities are given by a
softmax nonlinearity, the group of units $\v^{(i)}$ are referred to as
softmax units~\citep{Salakhutdinov-NIPS2010-softmax}.

\section{Difficulties with Word Observations}

While in the binary case the size of the visible layer is equal to data dimenionality, in the $K$-ary case the size of the visible
layer is $K$ times the dimensionality.  For language processing
applications, where~$K$ is the vocabulary size and can run into the
hundred thousands, the visible layer can become unmanageably
large.



The difficulty with large~$K$ is that the Gibbs operator on the
visible units becomes expensive to simulate, making it difficult to
perform updates of the negative samples.  That is, generating a sample
from the conditional distribution in Eq.~(\ref{eqn:karyviscond})
dominates the stochastic learning procedure as~$K$ increases.
The reason for this expense is that it is necessary to compute the
activity associated with \emph{each} of the $K$ possible outcomes, even though
only a \emph{single} one will actually be selected.

On the other hand, given a mini-batch $\{\v_t\}$ and negative samples
$\{\tilde{\v}_m\}$, the gradient computations in
Eq.~(\ref{eqn:approx_gradient}) are able to take advantage of the
sparsity of the visible activity.  Since each~$\v_t$
and~$\tilde{\v}_m$ only contain~$n$ non-zero entries, the cost of
the gradient estimator has no dependence on~$K$ and can be rapidly
computed.  Thus the only barrier to efficient learning of
high-dimensional multinomial RBMs is the complexity of the Gibbs
update for the visible units.


Dealing with large multinomial distributions is an issue that has come
up previously in work on neural network language models
\citep{nnlm:2001:nips}. For example, \citet{Morin+al-2005} addressed
this problem by introducing a fixed factorization of the (conditional)
multinomial using a binary tree in which each leaf is associated with
a single word.  The tree was determined using an external knowledge
base, although \citet{Mnih+Hinton-2009} investigated ways of extending
this approach by learning the word tree from data.

Unfortunately, tree-structured solutions are not applicable to the
problem of modeling the \emph{joint} distribution of $n$ consecutive
words, as we wish to do here. Introducing a directed tree breaks the
undirected, symmetric nature of the interaction between the visible
and hidden units of the RBM.  While one strategy might be to use a
conditional RBM to model the tree-based factorizations, similar to 
\citet{Mnih+Hinton-2009}, the end result would not be an RBM
model of $n$-gram word windows, nor would it even be a conditional RBM
over the next word given the $n-1$ previous ones.

In summary, dealing with $K$-ary observations in the Boltzmann machine
framework for large $K$ is a crucial open problem that has inhibited
the development of deep learning solutions NLP problems.

\section{Metropolis--Hastings for Softmax Units}

Having identified the Gibbs update of the visible units as the
limiting factor in efficient learning of large-$K$ multinomial
observations, it is natural to examine whether other operators might
be used for the Monte Carlo estimate in
Eq.~(\ref{eqn:approx_gradient}).  In particular, we desire a
transition operator that can take advantage of the same sparse
operations that enable the gradient to be efficiently computed from
the positive and negative samples, while still leaving~$p(\v,\h)$
invariant and thus still satisfying the convergence conditions of the stochastic approximation learning
procedure.

To achieve this, instead of sampling exactly from the conditionals
$p(\v^{(i)}|\h)$ within the Markov chain, we use a small number of
iterations of Metropolis--Hastings (M--H) sampling.
Let~$q(\hat{\v}^{(i)} \gets \v^{(i)})$ be a proposal distribution for
group~$i$.  The following stochastic operator leaves~$p(\v,\h)$ invariant:
\begin{enumerate}
\item Given the current visible state~$\v$, sample a
  proposal~$\hat{\v}$ for group~$i$, such that~${\hat{\v}^{(i)} \sim
    q(\hat{\v}^{(i)} \gets \v^{(i)})}$ and~${\hat{\v}^{(j)}=\v^{(j)}}$
  for~${i\neq j}$ (i.e.\ sample a proposed new word for position $i$).
\item Replace the $i$th part of the current state $\v^{(i)}$ with
  $\hat{\v}^{(i)}$ with probability:
\end{enumerate}
\vspace{-0.5cm}%
  \begin{align*}
    \min\left\{1,\frac{q(\v^{(i)}\gets\hat{\v}^{(i)})
      \exp(\b^{(i)^\top} \hat{\v}^{(i)} + \h^\top \W^{(i)} \hat{\v}^{(i)})
    }{
      q(\hat{\v}^{(i)}\gets \v^{(i)}) \exp(\b^{(i)^\top} \v^{(i)} + \h^\top \W^{(i)} \v^{(i)})
    }\right\}.
  \end{align*}
Assuming it is possible to efficiently sample from the proposal
distribution~~${q(\hat{\v}^{(i)} \gets \v^{(i)})}$, this M--H operator
is fast to compute as it does not require normalizing over all possible values of the visible units in group~$i$ and, in fact, only requires the unnormalized probability of one of them. Moreover, as the~$n$ visible groups are conditionally
independent given the hiddens, each group can
be simulated in parallel (i.e., words are sampled at every position separately).  The efficiency of these operations make it
possible to apply this transition operator many times before moving on
to other parts of the learning and still obtain a large speedup over exact sampling from the conditional.

\subsection{Efficient Sampling of Proposed Words}
\label{sect:alias_method}
The utility of M--H sampling for an RBM with word observations relies
on the fact that sampling from the proposal
$q(\hat{\v}^{(i)}\!\gets\!\v^{(i)})$ is much more efficient than sampling
from the correct softmax multinomial.  Although there are many
possibilities for designing such proposals, here we will explore the
most basic variant: \emph{independence chain} Metropolis--Hastings in
which the proposal distribution is fixed to be the marginal
distribution over words in the corpus.

Na\"{i}ve procedures for sampling from discrete distributions typically
have linear time complexity in the number of outcomes.  However, the
\emph{alias method} (pseudocode at \url{www.cs.toronto.edu/~gdahl}) of \citet{KronmalR1979} can be used to generate
samples in constant time with linear setup time.  While the
alias method would not help us construct a Gibbs sampler for the
visibles, it does make it possible to generate proposals extremely
efficiently, which we can then use to simulate the
Metropolis--Hastings operator, regardless of the current target distribution.

The alias method leverages the fact that any $K$-valued discrete
distribution can be written as a uniform mixture of $K$ Bernoulli
distributions.  Having constructed this mixture distribution at setup
time (with linear time and space cost), new samples can be generated
in constant time by sampling uniformly from the~$K$ mixture components, followed by sampling from that component's Bernoulli distribution.

\begin{figure}[t]
  \centering%
  \subfigure[Example KL Convergence]{%
    \centering%
    \includegraphics[width=0.23\textwidth]%
    {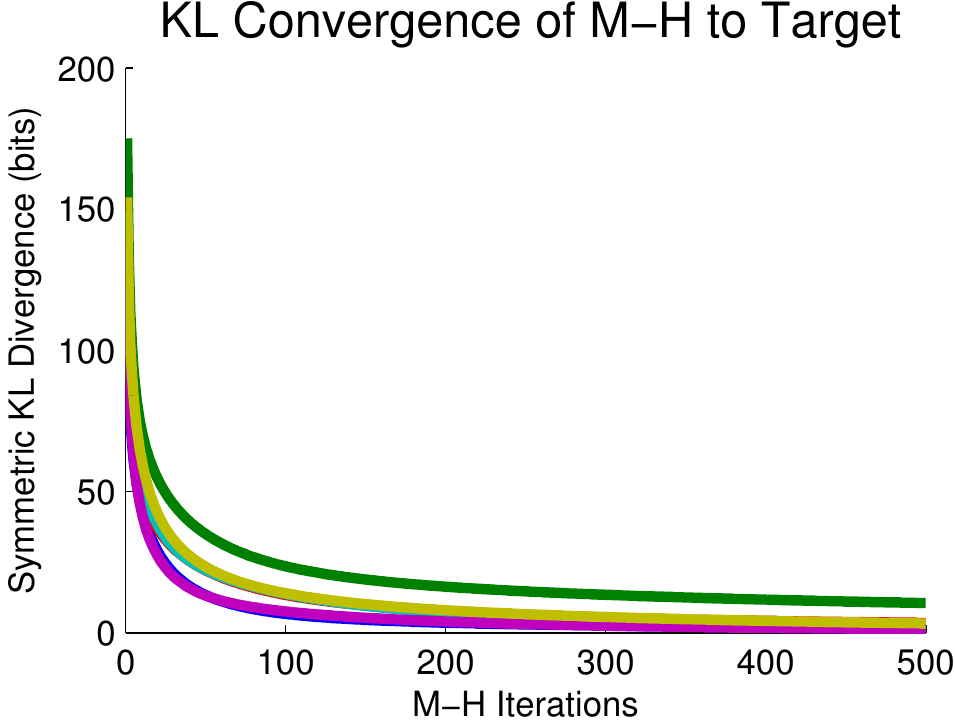}%
    \label{fig:kl-agg}%
  }~%
  \subfigure[Example TV Convergence]{%
    \centering%
    \includegraphics[width=0.23\textwidth]%
    {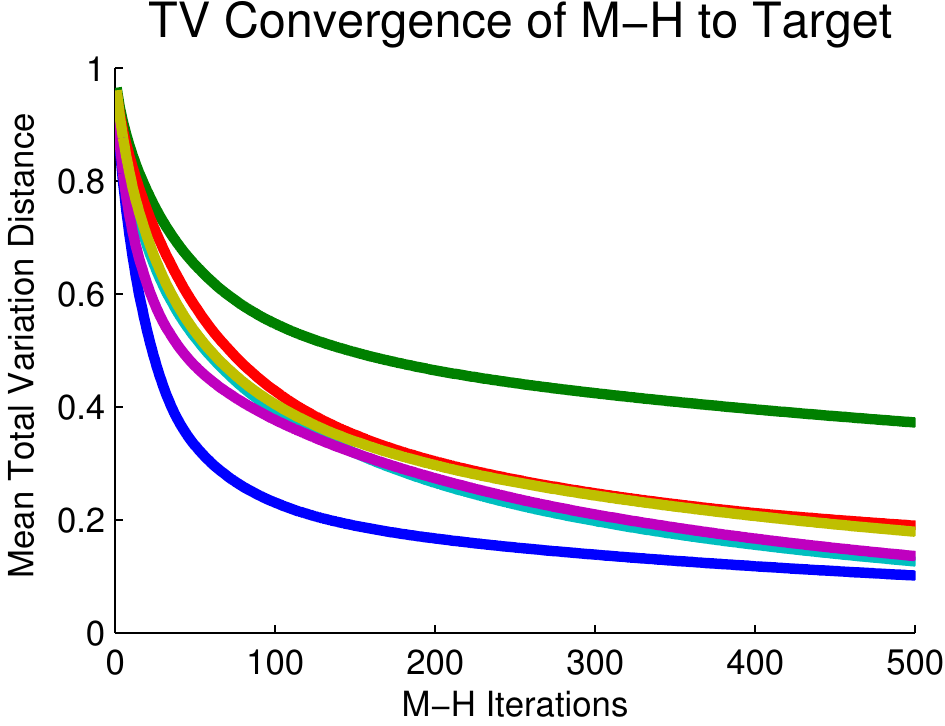}%
    \label{fig:tv-agg}%
  }\\%
  \vspace{-0.25cm}%
  \subfigure[Example Per-Word KL]{%
    \centering%
    \includegraphics[width=0.23\textwidth]%
    {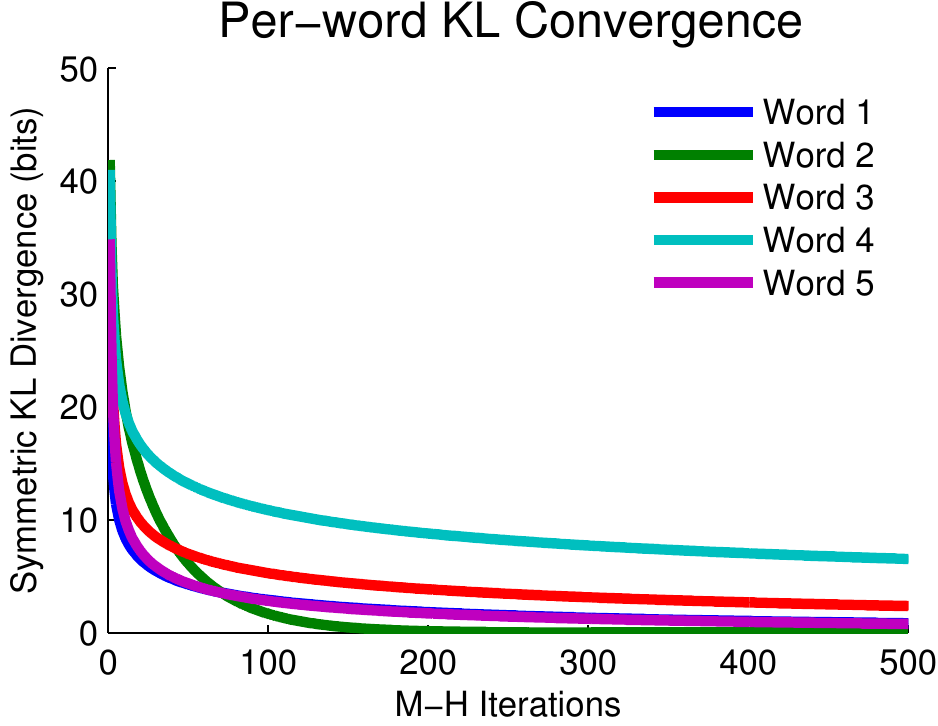}%
    \label{fig:kl-single}%
  }~%
  \subfigure[Example Per-Word TV]{%
    \centering%
    \includegraphics[width=0.23\textwidth]%
    {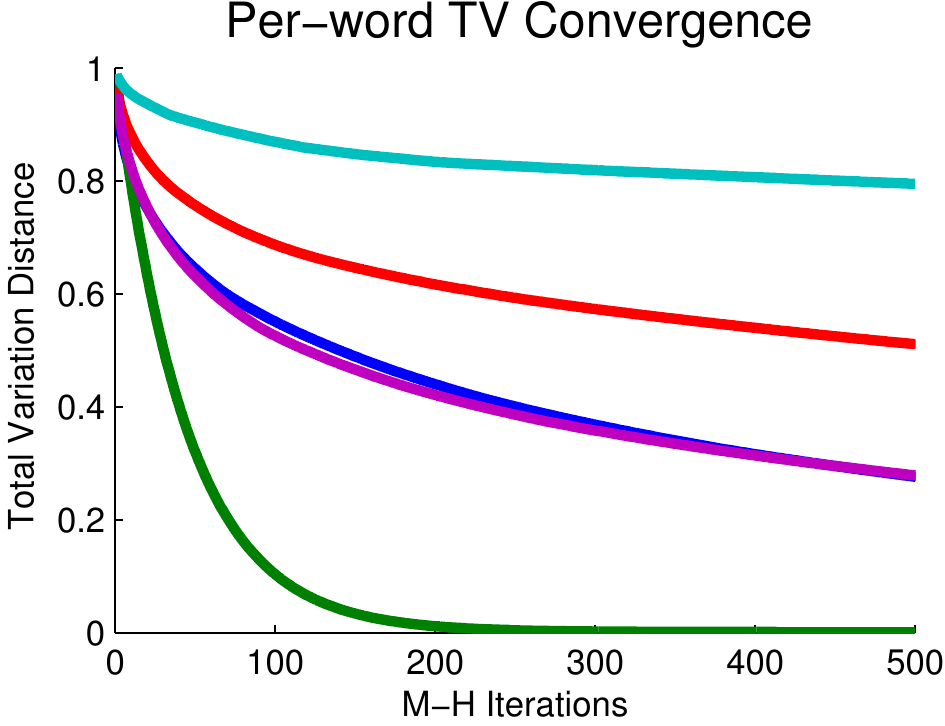}%
    \label{fig:tv-single}%
  }%
  \vspace{-0.4cm}%
  \caption{Convergence of the Metropolis--Hastings operator to the
    true conditional distribution over the visibles for a trained
    5-gram RBM with a vocabulary of 100K words.
    (a)~KL divergence for six
    randomly-chosen data cases.  (b)~Average total variation distance
    for the same six cases.  (c,d)~For the slowest-converging
    of the six top curves (dark green), these are broken down for each
    of the five multinomials in KL and total variation, respectively.}
\vspace*{-0.6cm}
\end{figure}

\subsection{Mixing of Metropolis--Hastings}
Although this procedure eliminates dependence of the learning
algorithm on~$K$, it is important to examine the mixing of
Metropolis--Hastings and how sensitive it is to~$K$ in
practice. Although there is evidence \cite{hinton2002} that
poorly-mixing Markov chains can yield good learning signals, when this
will occur is not as well understood.  We examined the mixing issue
using the model described in Section~\ref{sect:ngram_rbm} with the
parameters learned from the Gigaword corpus with a 100,000-word
vocabulary as described in Section~\ref{sec:chunking}.

We analytically computed the distributions implied by 
iterations of the M--H operator, assuming the initial state was
drawn according to~$\prod_i q(\v^{(i)})$.  As this computation
requires the instantiation of~$n$ ${100\sf{k}\times 100\sf{k}}$
matrices, it cannot be done at training time, but was done offline for
analysis purposes.  Each application of Metropolis--Hastings results
in a new distribution converging to the target (true) conditional.

Figures~\ref{fig:kl-agg} and~\ref{fig:tv-agg} show this convergence
for the ``reconstruction'' distributions of six randomly-chosen
5-grams from the corpus, using two metrics: symmetric
Kullback--Leibler (KL) divergence and total variation (TV) distance,
which is the standard measure for analysis of MCMC mixing.  The TV
distance shown is the mean across the five group distributions.
Figures~\ref{fig:kl-single} and~\ref{fig:tv-single} show these metrics
broken down by grouping, for the slowest curves (dark green) of the
top two figures.  These curves highlight that the state of the hidden
units has a strong impact on the mixing and that most groups mix very
quickly while a few converge slowly.  We feel that these curves, along
with the results of Section~\ref{sect:appl}, indicate that
the mixing is effective, but could benefit from further study.

\section{Related Work}

Using M--H sampling for a multinomial distribution with softmax
probabilities has been explored in the context of a neural network
language model by \citet{Bengio+Senecal-2003}. They used M--H to
estimate the training gradient at the output of the neural network.
However, their work did not address or investigate its use in the
context of Boltzmann machines in general.

\citet{Salakhutdinov-NIPS2010-softmax} describe an alternative to
directed topic models called the \emph{replicated softmax} RBM that
uses softmax units over the entire vocabulary with tied weights to
model an \emph{unordered} collection of words (a.k.a.\ bag of
words). Since their RBM ties the weights to all the words in a single
document, there is only one vocabulary-sized multinomial distribution
to compute per document, instead of the $n$ required when modeling a
window of consecutive words.  Therefore sampling a document
conditioned on the hidden variables of the replicated softmax still
incurs a computational cost linear in $K$, although the problem is not
amplified by a factor of $n$ as it is here.  Notably,
\citet{Salakhutdinov-NIPS2010-softmax} limited their vocabulary
to~${K<14,000}$.



No known previous work has attempted to address the computational
burden associated with $K$-ary observations with large $K$ in
RBMs. The M--H-based approach used here is not specific to a particular
Boltzmann machine and could be used for any model with large softmax
units, although the applications that motivate us come from
NLP. Dealing with the large softmax problem is essential if Boltzmann
machines are to be practical for natural language data.

In Section~\ref{sect:appl}, we present results on the task of learning
word representations.  This task has been
investigated previously by others. \citet{Turian+Ratinov+Bengio-2010}
provide an overview and evaluation of these different methods,
including those of \citet{Mnih+Hinton-2009} and of
\citet{CollobertR2008}.  We have already mentioned the work of
\citet{Mnih+Hinton-2009}, who model the conditional distribution of
the last word in $n$-gram windows. \citet{CollobertR2008} follows a
similar approach, by training a neural network to fill-in the middle
word of an $n$-gram window, using a margin-based learning objective.
In contrast, we model the \emph{joint} distribution of the whole
$n$-gram window, which implies that the RBM could be used to fill-in
any word within a window. Moreover, inference with an RBM yields a
hidden representation of the whole window and not simply of a single
word.

\section{Experiments}
\label{sect:appl}

We evaluated our M--H approach to
training RBMs on two NLP tasks: chunking and sentiment classification.
Both applications will be based on the same RBM model over $n$-gram
windows of words, hence we first describe the parameterization of this
RBM and later present how it was used for chunking and sentiment
classification. Both applications also take advantage of the model's
ability to learn useful feature vectors for entire $n$-grams, not just
individual words.

\subsection{RBM Model of $n$-gram Windows}
\label{sect:ngram_rbm}

In the standard parameterization presented in Section~\ref{sec:rbm},
the RBM uses separate weights (i.e., different columns of $\W$) to
model observations at different positions. When
training an RBM on word $n$-gram windows, we would prefer to share
parameters across identical words in different positions in the
window and factor the weights into position-dependent weights and
position-independent weights (word representations).

Therefore, we use an RBM parameterization
very similar to that of \citet{Mnih+Hinton-2007}, which itself is
inspired by previous work on neural language models~\citep{nnlm:2001:nips}.
The idea is to learn, for each possible word $w$, a lower-dimensional
linear projection of its one-hot encoding by incorporating the
projection directly in the energy function of the RBM. Moreover, we
share this projection across positions within the $n$-gram window.
Let~$\D$ be the matrix of this linear projection and let $\e_w$ be the
one-hot representation of $w$ (where we treat $w$ as an integer index
in the vocabulary), performing this projection $\D \e_w$ is equivalent
to selecting the appropriate column $\D_{\cdot,w}$ of this matrix.
This column vector can then be seen as a real-valued vector
representation of that word. The real-valued vector representations of
all words within the $n$-gram are then concatenated and connected to
the hidden layer with a single weight matrix.

More specifically, let $\D$ be the ${D \times K}$ matrix of word
representations. These word representations are introduced by reparameterizing
${\W^{(i)} = \U^{(i)} \D}$, where $\U^{(i)}$ is a position-dependent ${H \times
D}$ matrix.  The biases across positions are also shared, i.e., we
learn a single bias vector $\b^*$ that is used at all positions ($\b^{(i)}
= \b^*~\forall i$). The energy function becomes
\begin{align*}
  E(\v,\h) = - \c^\top \h + \sum_{i=1}^n -\b^{*^\top} \v^{(i)}  - \h^\top ~\U^{(i)} ~\D \v^{(i)}
\end{align*}
with conditional distributions
\begin{flalign*}
  p(\h|\v) & = \prod_j p(h_j| \v) &
  p(\v|\h) & = \prod_{i=1}^n p(\v^{(i)} | \h)
\end{flalign*}
\vspace{-0.4cm}%
\begin{align*}
  p(h_j=1|\v) & = \sigm\left(c_j + \sum_{i=1}^n \U_{j \cdot}^{(i)}~\D \v^{(i)}\right) \\
  p(\v^{(i)} = \e_k |\h) & = \frac{\exp(\b^{*^\top} \e_k + \h^\top \U^{(i)}~\D \e_k)}{\sum_{k'=1}^K \exp(\b^{*^\top} \e_{k'} + \h^\top \U^{(i)}~\D \e_{k'})}
\end{align*}
where $\U_{j \cdot}^{(i)}$ refers to the $j^{\rm th}$ row vector of
$\U^{(i)}$. The gradients with respect to this parameterization are easily
derived from Eq.~(\ref{eqn:approx_gradient}).
We refer to this construction as a \emph{word
representation RBM} {(WRRBM)}.  

In contrast to \citet{Mnih+Hinton-2007}, rather than training the
WRRBM conditionally to model~$p(w_{n+t-1}|w_{t},\dots,w_{n+t-2})$, we
train it using Metropolis--Hastings to model the full joint
distribution $p(w_t,\dots,w_{n+t-1})$.  That is, we train the WRRBM
based on the objective
\begin{align*}
  {\cal L(\theta)} =\!-\! \sum_{t}\! \log p(\v^{(1)}\!=\!\e_{w_t},\v^{(2)}\!=\!\e_{w_{t+1}},\dots,\v^{(n)}_{w_{n+t-1}})
\end{align*}
using stochastic approximation from M--H sampling of the word
observations.  For models with~${n>2}$, we also found it helpful to
incorporate~$\ell_2$ regularization of the weights, and to use
momentum when updating $\U^{(i)}$.

\subsection{Chunking Task}
\label{sec:chunking}

As described by \citet{Turian+Ratinov+Bengio-2010}, learning
real-valued word representations can be used as a simple way of
performing semi-supervised learning for a given method, by first
learning word representations on unlabeled text and then feeding these
representations as additional features to a supervised learning model.

We trained the WRRBM on windows of text derived from the English
Gigaword
corpus\footnote{\url{http://www.ldc.upenn.edu/Catalog/catalogEntry.jsp?catalogId=LDC2005T12}}.
The dataset is a corpus of newswire text from a variety of sources. We
extracted each news story and trained only on windows of $n$ words
that did not cross the boundary between two different stories. We used
NLTK \citep{BirdKleinLoper09} to tokenize the words and sentences, and
also corrected a few common punctuation-related tokenization errors.
As in \citet{collobert:2011b}, we lowercased all words and
delexicalized numbers (replacing consecutive occurrences of one or
more digits inside a word with just a single \# character).  Unlike
\citet{collobert:2011b}, we did not include additional capitalization
features, but discarded all capitalization information.  We used a
vocabulary consisting of the 100,000 most frequent words plus a
special ``unknown word'' token to which all remaining words were
mapped.

We evaluated the learned WRRBM word representations on a chunking
task, following the setup described in
\citet{Turian+Ratinov+Bengio-2010} and using the associated
publicly-available code, as well as
CRFSuite\footnote{\url{http://www.chokkan.org/software/crfsuite/}}. As
in \citet{Turian+Ratinov+Bengio-2010}, we used data from the
CoNLL-2000 shared task.  We used a scale of $0.1$ for the word
representation features (as \citet{Turian+Ratinov+Bengio-2010}
recommend) and for each WRRBM model, tried~$\ell_2$ penalties ${\lambda \in
\{0.0001, 1.2, 2.4, 3.2\}}$ for CRF training. We selected the single
model with the best validation F1 score over all runs and evaluated it
on the test set. The model with the best validation F1 score used
$3$-gram word windows, ${\lambda=1.2}$, 250 hidden units, a learning
rate of 0.01, and used 100 steps of M--H sampling to update each word
observation in the negative data.

The results are reported in Table~\ref{tab:results-chunking}, where we
observe that word representations learned by our model achieved higher
validation and test scores than the baseline of not using word
representation features, and are comparable to the best of the three
word representations tried in
\citet{Turian+Ratinov+Bengio-2010}\footnote{Better results have been
  reported by others for this dataset: the spectral approach of \citet{DhillonP2011} used different (less stringent) preprocessing and a vocabulary of
  300,000 words and obtained higher F1 scores than the methods
  evaluated in \citet{Turian+Ratinov+Bengio-2010}.  Unfortunately, the vocabulary and preprocessing differences mean that neither our result nor the one in \citet{Turian+Ratinov+Bengio-2010} are directly comparable to \citet{DhillonP2011}.
 }.

Although the word representations learned by our model are highly
effective features for chunking, an important advantage of our model
over many other ways of inducing word representations is that it also
naturally produces a feature vector for the entire $n$-gram. For the
trigram model mentioned above, we also tried adding the hidden unit
activation probability vector as a feature for chunking. For each word
$w_i$ in the input sentence, we generated features using the hidden
unit activation probabilities for the trigram $w_{i-1} w_{i}
w_{i+1}$. No features were generated for the first and last word of
the sentence. The hidden unit activation probability features
improved validation set F1 to 95.01 and test set F1 to 94.44, a test
set result superior to all word embedding results on chunking reported
in \citet{Turian+Ratinov+Bengio-2010}.

\begin{table}
  \vspace*{-0.25cm}%
  \caption{Comparison of experimental results on the chunking task.
    The baseline results were taken from
    \citet{Turian+Ratinov+Bengio-2010}. The performance measure is
    F1.}%
  \label{tab:results-chunking}
  \vspace*{.1in}
  \begin{tabular}[b]{l|c|c}
    Model & Valid & Test \\ \hline \hline
    CRF w/o word representations & 94.16 & 93.79\\
    HLBL {\small \citep{Mnih+Hinton-2009}} & 94.63 & 94.00\\
    C\&W {\small \citep{CollobertR2008}}& 94.66 & 94.10\\
    Brown clusters & 94.67 & 94.11\\ \hline
    WRRBM & 94.82 & 94.10 \\
    WRRBM (with hidden units) & {\bf 95.01} & {\bf 94.44}
 \end{tabular}
\vspace*{-0.8cm}%
\end{table}


As can be seen in Table~\ref{wordNeighbors}, the learned word
representations capture meaningful information about words.
However, the model primarily learns word representations that
capture syntactic information (as do the representations studied in
\citet{Turian+Ratinov+Bengio-2010}), as it only models short
windows of text and must enforce local agreement. 
Nevertheless, word representations capture some semantic information,
but only after similar syntactic roles have been enforced.  Although
not shown in Table~\ref{wordNeighbors}, the model consistently embeds
the following natural groups of words together (maintaining small
intra-group distances): days of the week, words for single digit
numbers, months of the year, and abbreviations for months of the
year. A 2D visualization of the word representations generated by
t-SNE~\citep{VanDerMaaten08} is provided at
\url{http://i.imgur.com/ZbrzO.png}.

\subsection{Sentiment Classification Task}

\citet{Maas-etal2011} describe a model designed to learn word
representations specifically for sentiment analysis. They train a
probabilistic model of documents that is capable of learning word
representations and leveraging sentiment labels in a semi-supervised
framework. Even without using the sentiment labels, by treating each
document as a single bag of words, their model tends to learn
distributed representations for words that capture mostly semantic
information since the co-occurrence of words in documents encodes very
little syntactic information. To get the best results on sentiment
classification, they combined features learned by their model with 
bag-of-words feature vectors (normalized to unit length) using binary term
frequency weights (referred to as ``bnc'').

We applied the WRRBM to the problem of sentiment classification by
treating a document as a ``bag of $n$-grams'', as this maps well
onto the fixed-window model for text.  At first glance, a word
representation RBM might not seem to be a suitable model for
learning features to improve sentiment classification. A WRRBM trained
on the phrases ``this movie is wonderful'' and ``this movie is
atrocious'' will learn that the word ``wonderful'' and the word
``atrocious'' can appear in similar contexts and thus should have
vectors near each other, even though they should be treated very
differently for sentiment analysis.  However, a class-conditional
model that trains separate WRRBMs on $n$-grams from documents
expressing positive and negative sentiment avoids this problem.

We trained class-specific, 5-gram WRRBMs on the labeled documents of
the Large Movie Review dataset introduced by \citet{Maas-etal2011},
independently parameterizing words that occurred at least 235 times in
the training set (giving us approximately the same vocabulary size as
the model used in \citet{Maas-etal2011}). 

To label a test document using the class-specific WRRBM, we fit a
threshold to the difference between the average free energies assigned
to $n$-grams in the document by the positive-sentiment and negative
sentiment models.  We explored a variety of different hyperparameters
(number of hidden units, training parameters, and $n$) for the pairs
of WRRBMs and selected the WRRBM pair giving best training set
classification performance. This WRRBM pair yielded 87.42\% accuracy
on the test set.

We additionally examined the performance gain by appending to the
bag-of-words features the average $n$-gram free energies under
both class-specific WRRBMs.  The bag-of-words feature vector was
weighted and normalized as in \citet{Maas-etal2011} and the average
free energies were scaled to lie on~$[0,1]$.  We then trained a linear
SVM to classify documents based on the resulting document feature
vectors, giving us $89.23$\% accuracy on the test set. This result is
the best known result on this benchmark and, notably, our method did not
make use of the unlabeled data.

\begin{table}
  \caption{Experimental results on the sentiment classification task. The baseline results were taken
    from \citet{Maas-etal2011}. The performance measure is accuracy (\%).}%
  \label{tab:results-sentiment}%
\vspace*{.1in}
  \begin{tabular}[b]{l|c}
    Model & Test \\ \hline \hline
    LDA & 67.42 \\
    LSA & 83.96 \\
    \citet{Maas-etal2011}'s ``full'' method & 87.44 \\
    Bag of words ``bnc'' & 87.80 \\
    \citet{Maas-etal2011}'s ``full'' method & \multirow{2}{*}{88.33} \\
    ~~+ bag of words ``bnc'' & \\ 
    \citet{Maas-etal2011}'s ``full'' method & \multirow{2}{*}{88.89} \\
    ~~+ bag of words ``bnc'' + unlabeled data & \\ \hline
    WRRBM & 87.42 \\
    WRRBM + bag of words ``bnc'' & {\bf 89.23} \\
 \end{tabular}
\vspace{-0.5cm}
\end{table}

\begin{table*}
\caption{The five nearest neighbors (in the word feature vector space) of some sample words. 
}
\label{wordNeighbors}
\vspace*{.2cm}
\begin{center}
\begin{tabular}{|l|l|l|l|l|l|l|}
\hline
{\bf could} & {\bf spokeswoman } & {\bf suspects } & {\bf science } & {\bf china } & {\bf mother } & {\bf sunday}\\
\hline
should & spokesman & defendants & sciences & japan & father & saturday\\
would & lawyer & detainees & medicine & taiwan & daughter & friday\\
will & columnist & hijackers & research & thailand & son & monday\\
can & consultant & attackers & economics & russia & grandmother & thursday\\
might & secretary-general & demonstrators & engineering & indonesia & sister & wednesday\\
\hline \hline
{\bf tom } & {\bf actually } & {\bf probably } & {\bf quickly } & {\bf earned } & {\bf what } & {\bf hotel}\\
\hline
jim & finally & certainly & easily & averaged & why & restaurant\\
bob & definitely & definitely & slowly & clinched & how & theater\\
kevin & rarely & hardly & carefully & retained & whether & casino\\
brian & eventually & usually & effectively & regained & whatever & ranch\\
steve & hardly & actually & frequently & grabbed & where & zoo\\
\hline
\end{tabular}
\end{center}
\end{table*}

\section{Conclusion}

We have described a method for training RBMs with large $K$-ary softmax units that results in weight updates with a computational cost independent of $K$, allowing for efficient learning even when $K$ is large. Using our method, we were able to train RBMs that learn meaningful representations of words and $n$-grams. Our results demonstrated the benefits of these features for chunking and sentiment classification and, given these successes, we are eager to try RBM-based models on other NLP tasks. Although the simple proposal distribution we used for M-H updates in this work is effective, exploring more sophisticated proposal distributions is an exciting prospect for future work.


\bibliography{mendeley}

\begin{thebibliography}{21}
\providecommand{\natexlab}[1]{#1}
\providecommand{\url}[1]{\texttt{#1}}
\expandafter\ifx\csname urlstyle\endcsname\relax
  \providecommand{\doi}[1]{doi: #1}\else
  \providecommand{\doi}{doi: \begingroup \urlstyle{rm}\Url}\fi

\bibitem[Smolensky(1986)]{Smolensky86}
P.~Smolensky.
\newblock {Information Processing in Dynamical Systems: Foundations of Harmony
  Theory}.
\newblock In \emph{Parallel Distributed Processing: Explorations in the
  Microstructure of Cognition}, pages 194--281. MIT Press, Cambridge, 1986.

\bibitem[Freund and Haussler(1991)]{FreundY1991}
Y.~Freund and D.~Haussler.
\newblock {Unsupervised learning of distributions of binary vectors using
  2-layer networks}.
\newblock In \emph{NIPS 4}, pages 912--919, 1991.

\bibitem[Ranzato et~al.(2010)Ranzato, Krizhevsky, and
  Hinton]{ranzato2010factored}
M.~Ranzato, A.~Krizhevsky, and G.~E. Hinton.
\newblock {{Factored 3-way restricted Boltzmann} machines for modeling natural
  images}.
\newblock In \emph{AISTATS}, 2010.

\bibitem[Salakhutdinov and Hinton(2009)]{Salakhutdinov-NIPS2010-softmax}
R.~Salakhutdinov and G.~E. Hinton.
\newblock {Replicated Softmax: an Undirected Topic Model}.
\newblock In \emph{NIPS 22}, pages 1607--1614, 2009.

\bibitem[Salakhutdinov et~al.(2007)Salakhutdinov, Mnih, and
  Hinton]{SalakhutdinovR2007b-small}
R.~Salakhutdinov, A.~Mnih, and G.~E. Hinton.
\newblock {Restricted {Boltzmann} machines for collaborative filtering}.
\newblock In \emph{ICML}, 2007.

\bibitem[Welling et~al.(2005)Welling, Rosen-Zvi, and
  Hinton]{WellingNIPS17-small}
M.~Welling, M.~Rosen-Zvi, and G.~E. Hinton.
\newblock Exponential family harmoniums with an application to information
  retrieval.
\newblock In \emph{NIPS 17}, 2005.

\bibitem[Younes(1989)]{YounesL1989}
L.~Younes.
\newblock {Parameter inference for imperfectly observed Gibbsian fields}.
\newblock \emph{Probability Theory Related Fields}, 82:\penalty0 625--645,
  1989.

\bibitem[Bengio et~al.(2001)Bengio, Ducharme, and Vincent]{nnlm:2001:nips}
Y.~Bengio, R.~Ducharme, and P.~Vincent.
\newblock A neural probabilistic language model.
\newblock In \emph{NIPS 13}, pages 932--938, 2001.

\bibitem[Morin and Bengio(2005)]{Morin+al-2005}
F.~Morin and Y.~Bengio.
\newblock Hierarchical probabilistic neural network language model.
\newblock In \emph{AISTATS}, 2005.

\bibitem[Mnih and Hinton(2009)]{Mnih+Hinton-2009}
A.~Mnih and G.~E. Hinton.
\newblock A scalable hierarchical distributed language model.
\newblock In \emph{NIPS 21}, pages 1081--1088, 2009.

\bibitem[Kronmal and Perterson(1979)]{KronmalR1979}
R.~A. Kronmal and A.~V. Perterson.
\newblock On the alias method for generating random variables from a discrete
  distribution.
\newblock \emph{The American Statistician}, 33\penalty0 (4):\penalty0 214--218,
  1979.

\bibitem[Hinton(2002)]{hinton2002}
G.~E. Hinton.
\newblock {Training products of experts by minimizing contrastive divergence}.
\newblock \emph{Neural Computation}, 14:\penalty0 1771--1800, 2002.

\bibitem[Bengio and S\'{e}n\'{e}cal(2003)]{Bengio+Senecal-2003}
Y.~Bengio and J.-S. S\'{e}n\'{e}cal.
\newblock Quick training of probabilistic neural nets by importance sampling.
\newblock In \emph{AISTATS}, 2003.

\bibitem[Turian et~al.(2010)Turian, Ratinov, and
  Bengio]{Turian+Ratinov+Bengio-2010}
J.~Turian, L.~Ratinov, and Y.~Bengio.
\newblock Word representations: A simple and general method for semi-supervised
  learning.
\newblock In \emph{ACL}, pages 384--394, 2010.

\bibitem[Collobert and Weston(2008)]{CollobertR2008}
R.~Collobert and J.~Weston.
\newblock A unified architecture for natural language processing: Deep neural
  networks with multitask learning.
\newblock In \emph{ICML}, 2008.

\bibitem[Mnih and Hinton(2007)]{Mnih+Hinton-2007}
A.~Mnih and G.~E. Hinton.
\newblock Three new graphical models for statistical language modelling.
\newblock In \emph{ICML}, pages 641--648, 2007.

\bibitem[Bird et~al.(2009)Bird, Klein, and Loper]{BirdKleinLoper09}
S.~Bird, E.~Klein, and E.~Loper.
\newblock \emph{Natural Language Processing with Python: Analyzing Text with
  the Natural Language Toolkit}.
\newblock O'Reilly, 2009.

\bibitem[Collobert et~al.(2011)Collobert, Weston, Bottou, Karlen, Kavukcuoglu,
  and Kuksa]{collobert:2011b}
R.~Collobert, J.~Weston, L.~Bottou, M.~Karlen, K.~Kavukcuoglu, and P.~Kuksa.
\newblock Natural language processing (almost) from scratch.
\newblock \emph{Journal of Machine Learning Research}, 2011.

\bibitem[Dhillon et~al.(2011)Dhillon, Foster, and Ungar]{DhillonP2011}
P.~Dhillon, D.~P. Foster, and L.~Ungar.
\newblock Multi-view learning of word embeddings via {CCA}.
\newblock In \emph{NIPS 24}, pages 199--207, 2011.

\bibitem[van~der Maaten and Hinton(2008)]{VanDerMaaten08}
L.~van~der Maaten and G.~E. Hinton.
\newblock Visualizing data using t-{SNE}.
\newblock \emph{Journal of Machine Learning Research}, 9:\penalty0 2579--2605,
  2008.

\bibitem[Maas et~al.(2011)Maas, Daly, Pham, Huang, Ng, and
  Potts]{Maas-etal2011}
A.~L. Maas, R.~E. Daly, P.~T. Pham, D.~Huang, A.~Y. Ng, and C.~Potts.
\newblock Learning word vectors for sentiment analysis.
\newblock In \emph{ACL}, pages 142--150, June 2011.

\end{thebibliography}

\bibliographystyle{unsrtnat}

\end{document}